# A Stitching Algorithm for Automated Surface Inspection of Rotationally Symmetric Components

Tobias Schlagenhauf[1], Tim Brander[1], Jürgen Fleischer[1]

[1]Karlsruhe Institute of Technology (KIT)

wbk-Institute of Production Science

Kaiserstraße 12, 76131 Karlsruhe, Germany

## Abstract

This paper provides a novel approach to stitching surface images of rotationally symmetric parts. It presents a process pipeline that uses a feature-based stitching approach to create a distortion-free and true-to-life image from a video file. The developed process thus enables, for example, condition monitoring without having to view many individual images. For validation purposes, this will be demonstrated in the paper using the concrete example of a worn ball screw drive spindle. The developed algorithm aims at reproducing the functional principle of a line scan camera system, whereby the physical measuring systems are replaced by a feature-based approach. For evaluation of the stitching algorithms, metrics are used, some of which have only been developed in this work or have been supplemented by test procedures already in use. The applicability of the developed algorithm is not only limited to machine tool spindles. Instead, the developed method allows a general approach to the surface inspection of various rotationally symmetric components and can therefore be used in a variety of industrial applications. Deep-learning-based detection Algorithms can easily be implemented to generate a complete pipeline for failure detection and condition monitoring on rotationally symmetric parts.

## 1. Introduction

A central aspect of effective industrial production is the availability of production facilities and the quality of the products manufactured with them. For the automated monitoring of machine tools, an increasing number of sensor systems are used. Visual approaches are particularly suitable for surface inspection of defects. With the help of machine learning, the image data can be quickly evaluated with regard to taught-in defects. (Kumar, 2008). For examination of rotationally symmetric components, such as ball screws, it is advisable to stitch together the individual images to form a complete image of the component surface instead of using a classifier for each individual image.

There are already a number of commercially used stitching tools, such as the panorama function in the digital camera (Xiong and Pulli, 2010). However, these classical approaches have problems when it comes to taking precise stitching images of cylindrical surfaces, especially when the images are extremely feature-poor due to homogeneous surfaces. This paper describes an approach to solving these problems when stitching low-featured, rotationally symmetric surfaces. The developed stitcher will be validated using the example of a ball screw drive.

The remainder of the paper is structured as follows. Section 2 reviews the current state of the art in the field of stitching. Section 3 presents our own approach. Section 4 shows the results, the selection process, and the evaluation metrics developed specifically for this purpose. The newly developed stitching approach is also applied to a concrete application. Section 5 presents a conclusion and ideas for future work.

## 2. Related Research

Image processing is already an important factor in the industry in the field of condition monitoring. The application for rotationally symmetric surfaces was presented in various publications. In some works, single close-up images are analyzed to predict the tool wear of a CNC lathe. Using extracted features from the rotated surface images, conclusions are drawn about the waviness of the grooves and thus the quality of the process (Dutta et al. 2016). In other variants, the surface roughness of a turned part is measured using backlighting (Balasundaram and Ratnam, 2014; Kumar and Ratnam, 2015). In the work by (Fernández-Robles et al. 2016), the tool quality of a cylindrical machining head is checked using a vision system. The tool edges are detected and checked for wear with the help of edge detection. In order to perform a complete inspection, the processing head is rotated for a total of 24 times by 15°. In (Schlagenhauf et al. 2019), the authors proposed a camera-based system integrated in a machine tool for the condition monitoring of defects on ball screw drive spindles.

In the previous approaches, single images were used to test for wear. This reaches its limits if the wear to be examined has progressed beyond the image boundary or if distortion due to the cylindrical shape leads to a lack of information. Therefore, in some cases, it is better to take a kind of panoramic image for the examination in a first step. Breitmeier Messtechnik GmbH (Beyerer et al. 2016) uses this method for the quality inspection of cylinder inner walls of combustion engines (Fig. 1). A line scan camera system is used which performs an azimuthal scanning movement with constant distance to the cylinder wall. With line scan cameras, high resolutions can be achieved and it is also easier to ensure a targeted and at the same time uniform illumination of the entire image. The camera can be synchronized by an incremental encoder that measures the rotation speed of the cylinder. Thus, one image line always corresponds to the same spatial displacement. However, there must be a way to precisely assign the speed to the images taken by the line scan camera. Retrofitting, for example for larger machine tools, is costly. When using an area sensor for the same applications, the captured images must be registered to each other afterwards in order to stitch them together.

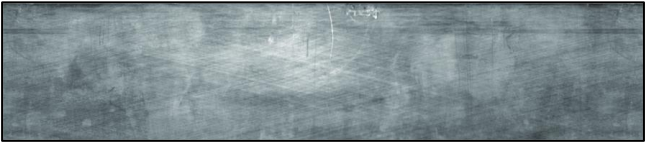

Fig. 1. Recording of the cylinder wall of a combustion engine by a line scan camera as used by Breitmeier Messtechnik GmbH (Beyerer et al. 2016)

## 2.1 Image Stitching Techniques

Extensive research literature and several commercial applications are available on stitching methods in image processing. Direct and feature-based techniques are considered the most important approaches for stitching images. The direct ones work by minimizing pixel-to-pixel dissimilarities, the feature-based techniques by extracting features, which are then matched.

### 2.1.1 Direct Techniques

In the direct techniques, all pixel intensities of the images are compared. Subsequently, the sum of the absolute difference between overlapping pixels is reduced. Other cost functions can also be used. The procedure is computationally complex since each pixel window is compared with the others. The main advantage of the direct method is that the information available for image alignment is used optimally. The main disadvantage, however, besides the computational effort, is the limited convergence range. (Adel et al. 2014; Brown and Lowe, 2007; Zitová and Flusser, 2003)

### 2.1.2 Feature-based Techniques

Most of the feature-based techniques can be roughly divided into five sub-steps. First, features are detected at distinctive points in the images and stored in a form which is easy to compare using a descriptor. Then, a list of corresponding feature points is generated with the help of a matcher. In the next step, these points serve as parameters for the calculation of a transformation matrix, which indicates the spatial displacement and distortion of image two with respect to image one.

To determine the coefficients of the transformation matrix, three non-collinear points are required for an affine image. These result in the following system of equations:

$$\begin{bmatrix} x'_1 & x'_2 & x'_3 \\ y'_1 & y'_2 & y'_3 \\ 1 & 1 & 1 \end{bmatrix} = \begin{bmatrix} a_{11} & a_{12} & t_x \\ a_{21} & a_{22} & t_y \\ 0 & 0 & 1 \end{bmatrix} \begin{bmatrix} x_1 & x_2 & x_3 \\ y_1 & y_2 & y_3 \\ 1 & 1 & 1 \end{bmatrix} \quad (1)$$

$$P' = AP \quad (2)$$

From this, $A$ is calculated to

$$A = P'P^{-1} \quad (3)$$

With more than three corresponding points, the parameters can be solved in the following system of equations:

$$A = P'P^T(PP^T)^{-1} \quad (4)$$

with

$$P'P^T = \begin{bmatrix} \sum x'_n x_n & \sum x'_n y_n & \sum x'_n \\ \sum y'_n x_n & \sum y'_n y_n & \sum y'_n \\ \sum x_n & \sum y_n & N \end{bmatrix} \quad (5)$$

$$PP^T = \begin{bmatrix} \sum x_n^2 & \sum x_n y_n & \sum x_n \\ \sum x_n y_n & \sum y_n^2 & \sum y_n \\ \sum x_n & \sum y_n & N \end{bmatrix} \quad (6)$$

(Jähne, 2012)

The fourth step consists in compositing, where emphasis is not only on choosing which method is used for bending and shifting, but also on determining on which surface the images are applied, for example on a flat, cylindrical, and spherical surface. The last step is blending during which edges or other artifacts caused by exposure and detection errors are compensated. Feature-based methods have the advantage of being potentially faster and more robust against scene motion. (Adel et al. 2014; Brown and Lowe, 2007; Tsen, 2014; Zitová and Flusser, 2003)

## 3. Own Approach

The process pipeline described in the introduction to the automatic surface inspection of rotationally symmetric components is shown in Fig. 2. The input consists of a video of the object to be examined. This is then combined into a kind of panoramic image and, based on this, a classification regarding defects is performed and evaluated. The stitching algorithm plays an important role in this process. It is validated in this paper using the example of a ball screw.

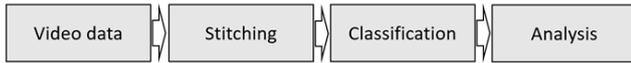
Fig. 2. Process pipeline for automatic surface inspection

## 3.1 Deficits of Classical Stitching Processes

Direct methods are used less and less for stitching tasks in current work due to poorer computational times and robustness as compared to feature-based variants. Also, it is easier to make incorrect assignments in low-feature environments because the pixel intensities are very similar there. (Adel et al. 2014; Zitová and Flusser, 2003)

The classical feature-based methods have particular problems when it comes to upscaling the process from a few frames to longer acquisition sequences. When only a few frames are merged, the natural approach is to select one of the frames as a reference and then transform all other subsequent frames to its reference coordinate system. This leads to results like those shown in Fig. 3 (left), since due to the cylindrical shape of the spindle and the perspective transformation between the individual frames, the image will fold further and further back. It is not possible either to transform the frames into a planar plane by an affine transformation in advance, since this distorts the images, whereby assignment of the errors found is only possible with errors (see Fig. 3 (right)).

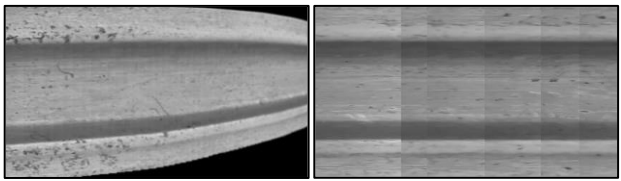
Fig. 3. Result of classical stitching with perspective transformation (left) and affine transformation (right) using the example of a tool spindle

## 3.2 New Stitching Method

The newly developed approach aims to make use of the functional principle of a line scan camera system. This system normally works with an incremental encoder. This sensor allows to calculate the displacement of the individual captured lines in relation to each other and, therefore, to create a realistic unrolled image even of cylindrical surfaces. The line spacing $x$ results from the relative speed $\dot{x}(t)$ of object and sensor and the line readout rate. The basic idea is to adapt the concept of the line scan camera and to replace the physical measuring systems by a feature-based approach. The high frame rate of current video recording systems favors this approach, since only a small shift occurs between the individual recordings (Fig. 5). Following the classical feature-based methods, the stitching algorithm is divided into the following steps: "preprocessing", "determination transformation model", "composition", and "blending" (Fig. 4).

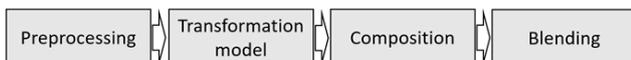
Fig. 4. Process steps of the stitching algorithm

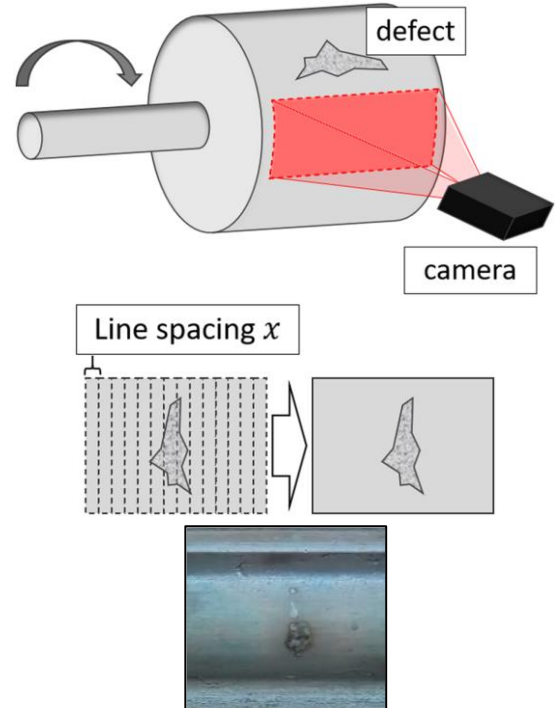
Fig. 5. Functional principle of the adapted line scan camera approach, and exemplary result from the ball screw drive – no perturbations are visible.

### 3.2.1 Preprocessing

Depending on the application, only a small section of the video recording is required for the analysis of existing defects. The larger section must therefore be cut to a narrower region of interest in a first step. Depending on the specific application, the input image must also be rotated. For the example of the spindle of a ball screw drive, first its thread pitch must be compensated and the thread shoulder of the selected region of interest must be rotated into the horizontal position. This makes it possible to inspect the complete component due to the translatory movement of the spindle under the camera. In a simplified application, it is also possible to examine a cylindrical body like a ball bearing or a drive shaft which is rotated only once around its own axis without translational movement. This eliminates the need to rotate the input images in the preprocessing step.

### 3.2.2 Determination of Transformation Model

This process step is exactly the same as in classical stitching. First, a detector analyzes the image coming from preprocessing and finds prominent feature points. Using a descriptor, the latter are converted into a form that is easier to distinguish and find again. Then, the process is repeated with the previous frame. To do this, a piece corresponding to the frame width is selected from the previously stitched panorama image. In the case of the first stitching iteration, the first frame is selected twice. Next, these points are put into a matcher, which creates a list of corresponding feature points by assigning them from the features of the image pair. To calculate the transformation matrix between both frames, this list is given in the last step in a resampling algorithm. This algorithm estimates a transformation model within a series of measured values in which outliers and gross errors also occur. This makes it very robust, should wrong features have been assigned to each other in the previous steps.

### 3.2.3 Composition

Instead of the classical approach, a new approach is chosen here. As already mentioned, the principle of a line scan camera system is reproduced. The procedure is shown in Fig. 6.

In a first step, the last complete frame (1) is taken from a previously stitched result image. This is now compared with the next frame (2) in the video. The comparison of both images is performed as described in the step "Determination of the transformation model" and finally delivers the transformation matrix (H) which distorts frame (2) perspectively and transfers it into the reference coordinate system of frame (1).

In the second step, this transformation matrix is applied to frame (2). The image is then displayed in relative displacement to frame (1) and its reference coordinate system. In classical variants, it would now be assembled with frame (1), but this would cause the result to fold back spatially into the image plane when scaling up to a large number of frames. This is due to the cylindrical surface and the resulting spatial distortion. Therefore, the new approach is based on determining only the displacement between the two frames in order to join the image sections together like an incremental encoder. The parameter of interest, as you can see in step two, is "col_min". After the image has been cropped by about 10% at the top and bottom, "col_min" is the first column in the image array where not a single black pixel with an intensity value of 0 is found. Therefore, "col_min" corresponds to the shift from frame (2) to frame (1) in horizontal direction. The cropping by 10% is necessary to cover certain outlier cases when feature detection is too poor and to continue to produce useful results.

The third step first extends the array from frame (1) by the shift "col_min" in horizontal direction and then inserts the unchanged frame (2) from the first step there. This ensures an undistorted and realistic image of the cylindrical surface, especially since the shift "col_min" is relatively small and only a few pixels wide due to the high frame rate in the video. This also makes it easier to ensure a targeted and at the same time uniform illumination of the entire image.

In the last step, the previous result array which is left after cutting out frame (1) in the first step is extended by the length of the stitched image from step three and finally the stitched array of frames (1) and (2) is inserted at the end.

The result is a panoramic image, which is extended by a new frame. This frame will be used again in the next iteration step as the already stitched result image, and the process starts again with the next video frame.

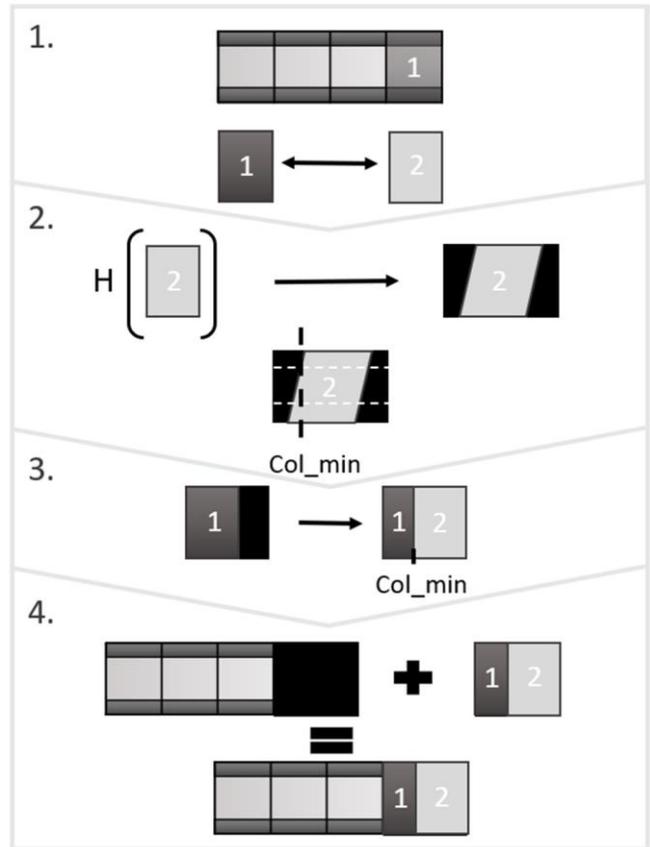

Fig. 6. Process flowchart for new composition approach

### 3.2.4 Blending

The stitching results from the previous steps still reveal edges (Fig. 7). This is due to slight differences in exposure between the individual frames. To correct this, a modification of the complex alpha blending is applied. A gradient image is used to create a smooth transition between two adjacent frames. On the left in the white area, the pixels each have a value of 255, which decreases linearly across the width of the image until they reach zero (black) at the right edge. A subrange of frame 1 is now multiplied by the gradient image; for frame 2, its inverse is used. This linear adjustment of the weighting results in a smooth transition. Applying it to only a certain part of the frames ensures that the areas outside are not unnecessarily often blended by several iterations, and, thus, the final result loses image sharpness. Choosing the best parameter results is a trade-off between an exposure fluctuation and image blur. The optimal parameters will be determined in the further course of the work. (see Fig. 8)

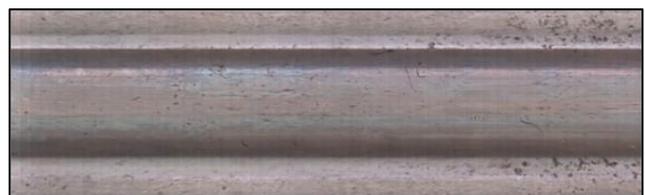

Fig. 7. Result with edges before blending

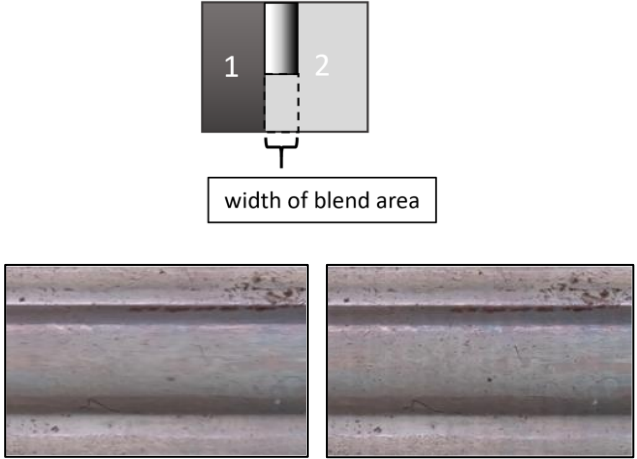

Fig. 8. Concept of complex alpha blending (top) and trade-off between loss of sharpness in result (left) and exposure fluctuation (right)

### 3.3 Classification

The next step in the analysis pipeline is a classification of the results regarding existing defects. As classification model, a VGG16 network is chosen. It has already proven its good performance in the benchmark test ImageNet Scale Visual Recognition (LSVRC) and is now quasi standard in many applications in the field of image classification. (Simonyan and Zisserman, 2014) Previously, a learning data set consisting of a total of 24739 samples was created and split into 80% for training and 20% for validation. The model is built and trained using the Keras library on a Nvidia DGX station.

The data input for the automatic classification is done by a sliding window approach, which divides the stitched image into predefined grids of size 150 x 150 pixels. These sections are classified subsequently. After assigning a patch to the class "defect", it is marked as a colored rectangle in the input image and the coordinates of the upper left point of the box are stored for further processing. Therewith, an exact localization of the defect on the component is possible. Further information such as the severity of the failure can be directly extracted from the image and used for further investigation.

For better documentation and quantification of the defect increase, the stitched image of the rotationally symmetric body is divided into ten areas. The spatial assignment of the detected defects is done by the coordinates of the upper left corner of the classified boundary boxes. The number of patches classified as wear per area is then stored in a JSON file for further data analysis.

## 4. Results

The code for the whole stitching algorithm can be found on our Github Repo. To create the process pipeline for the new feature-based stitcher, there are already a large number of different and proven algorithms for the respective "classical" sub-steps. A research delivered the following variants shown in Table 1. In their previous areas of application, these were characterized by robust, fast, and precise behavior, especially in homogeneous and low-contrast environments.

| Detector and Descriptor | |
|---|---|
| Oriented FAST and Rotated BRIEF (ORB) | (Adel et al. 2014) |
| Harris Corner (Harris Norm) | (Mistry, 2016) |
| Harris Subpixel (Harris Sub) | (Qiao et al. 2013) |
| Center-Surround Extrema (CenSurE) | (Gauglitz et al. 2011) |
| Matcher | |
| Nearest-Neighbor (BF) | (Brown and Lowe, 2007) |
| Nearest-Neighbor k-d (BF KNN) | (Brown and Lowe, 2007) |
| Fast Library for Approximate Nearest Neighbors (FLANN) | (Noble, 2016) |
| Resampling Algorithm | |
| Least Median of Squares (LMEDS) | (Massart et al. 1986) |
| Random Sample Consensus (RANSAC) | (Adel et al. 2014) |
| Progressive Sample Consensus (PROSAC) | Chum et al. 2005 |

Table. 1. Algorithm components

For evaluation of the stitching algorithms, metrics were used, some of which were only just developed or supplemented by test procedures already in use.

### 4.1 Evaluation Metrics

#### 4.1.1 Edge Metric

The edge metric compares the last pixel column of frame 1 with the first pixel column of frame 2 at the edge of two stitched frames. The hypothesis is that for an image without a visible edge, the difference between the exactly opposite pixel intensity values is minimally small. The clearer the visual edge, the greater the difference should be. For this purpose, the frames are converted into single-channel grayscale images whose pixel intensity is specified with a value between 0 and 255. By forming the absolute difference line by line, a value is obtained that is indicative of the correct alignment of two frames to each other:

$$EM = \frac{1}{n}\sum_{i=1}^{n}|x_{i1} - x_{i2}| \qquad (7)$$

with $x_{i1}$ as respective row value of the last column in frame 1 and $x_{i2}$ as respective row value of the first column in frame 2.

#### 4.1.2 Overlap Metric

This metric is generally based on the idea of the previous one, except that instead of individual pixel columns, whole areas are compared. These are the areas of the two frames that completely overlap. Since the two frames are overlapped when stitching, one of the two areas is not visible, but the more similar they are, e.g. in terms of exposure or correct alignment, the better the stitching result will be. To generate a comparison value, the absolute difference between the individual pixel values is calculated again, and a histogram is generated for each area and their differences are compared using various distance methods (Chi-squares, Euclidean, and Manhattan Distance). Another variant compares the Peak-Signal-Noise Ratio (PSNR) and is used in a similar form for quality comparison of compressed images. (Bind et al. 2013)

$$D_{area} = \frac{1}{n}\sum_{i=1}^{n}\sum_{j=1}^{n}|x_{ij1} - x_{ij2}| \quad (8)$$

$$PSNR = 20 * \log(\frac{255}{\sqrt{D_{area}}}) \quad (9)$$

$$OM = D_{area} + d_{euk} + d_{man} + d_{chi} - PSNR \quad (10)$$

The final equation of the overlap metric (OM) is composed of the absolute difference between the two areas $D_{area}$, the three distance methods ($d_{euk}, d_{man}, d_{chi}$) calculated from the histograms of both areas, and the Peak-Signal-Noise Ratio (PSNR) which includes $D_{area}$ as well.

### 4.1.3 Exposure Metric

With this metric, the pixel intensities are summed up for each column in an image, and a kind of histogram is created over the image width. If the image has strong and uneven variations between light and dark areas, this can be seen in the graph of which an example is shown in Fig.9. The bright areas are maxima, the dark are minima. The hypothesis is that the smaller the difference between the respective extremes, the smaller the fluctuations in brightness in the image under consideration. Since the large peaks of this curve are more meaningful and less subject to noise, only these peaks are considered. By calculating the difference from the local extremes, a value is obtained that expresses the exposure variation.

$$ExpM = \frac{1}{n}\sum_{i=1}^{n}|local_{max,i} - local_{min,i}| \quad (11)$$

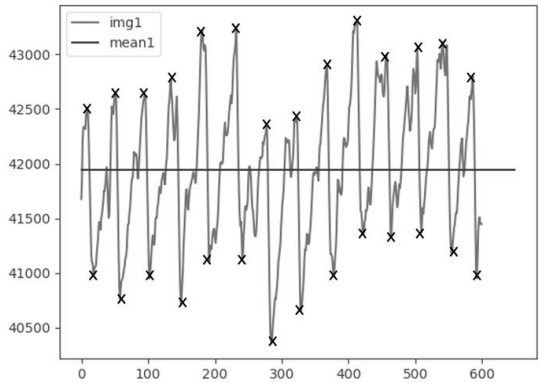

Fig. 9. Exemplary exposure graph and its local maxima and minima

### 4.1.4 Sharpness Metric

To determine the sharpness, a Fast Fourier Transformation (FFT) is applied. This is already used for automatic camera calibration. (Pertuez et al. 2013) On the other hand, the variance is calculated with a Laplace operator and is already used for the autofocusing of microscopes. (Pech-Pacheco et al. 2000)

### 4.2 Selection Process

The best combination of the different stitchers and the optimal blending parameters was determined in a four-phase test. A video section with few defects and a second one with many defects were used as input.

**Phase 1: Feature detection and matching**

In a first step, for each possible combination of feature detector, matcher, and resampling algorithm, the average number of features and matching numbers found during the stitching of 250 frames is measured. This part is essential, because the worse the results here are, the more error-prone the steps of the process pipeline based on them will be. Tests have shown that problems often occur when the number of matches found falls below ten. Therefore, in this first phase, it is also measured how often less than ten features and matches were found per stitching iteration and for visualization purposes shown in an overview table contained in our Gihub Repository[1].

**Phase 2: Edge and overlap metrics**

In the second phase, the remaining variants are tested for quality using the edge and overlap metrics and counting iteration steps with misleading alignment. From the respective measured values, a total sum is formed at the end. The PSNR values are subtracted from this, since the higher the PSNR values, the better the proven quality. Due to the distance metric to be minimized, the best variant therefore has the lowest total sum. The results for all combinations can again be found in our Github repository for visualization purposes.

**Phase 3: Processing time and subjective evaluation**

Now, a final selection is to be made from the four remaining candidates, for which, in addition to the temporal performance, a subjective evaluation is also to be included. The measured time indicates the time needed for each combination to stitch the 250 frames. As before, both a recording with few defects and one with many defects will be used (Fig. 10). Subsequently, an evaluation is made according to quality-related criteria, which are weighted differently depending on their influence. The best variant results from a combination of a Harris Corner detector, a brute force matcher, and a RANSAC resampling algorithm. It turns out that the more robust algorithms perform better. This is due to the fact that they cope better with the feature-poor recordings. Nevertheless, the adapted approach of the line scan camera makes it possible to produce high-quality results even without sub-pixel-precise feature detection.

---

[1] https://github.com/bra-ti/stitcher

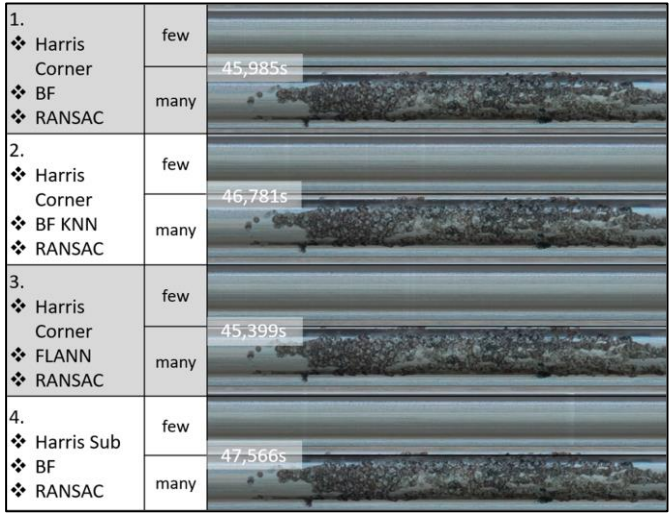

Fig. 10. Results of the four remaining combinations for feature-poor and feature-rich video sections and their corresponding processing times

**Phase 4: Blending parameters**

After the best variant has been identified, the blending parameter has to be determined. For this purpose, a test series with widths of 5, 25, 50, and 150 pixels is created. These correspond to 2.5%, 12.5%, 25%, and 75% of the frame width. The best width for the blending area WBP (Width Blending Parameter) also depends on how far the shift from one video frame to another is. If they are far apart, for example at a low video frame rate (frames per second), the larger the area that has to be blended, the greater the blending area. Therefore, the average pixel shift between the frames is also tracked. This finally determines a constant with which the best possible blending parameter can be calculated for each acquisition system.

Table 2 shows that for a transition width of 25 pixels, an optimum with low exposure value and high sharpness value is achieved. Since this value was achieved with an average pixel shift of 13.752 pixels (shift of the two frames to be stitched relative to each other), it is now necessary to find a generally valid statement for other recording settings. The following formula calculates a constant that gives the ratio of the best blending parameter to the measured pixel shift.

$$\beta = \frac{WBP}{\emptyset\ pixel\_shift} \quad (12)$$

After inserting the values, this results in the universally valid constant $\beta$ of 1.82. To determine the best blending parameter in still unknown videos, you can now simply apply formula (13).

$$WBP_{new} = 1{,}82 * \emptyset\ pixel\_shift \quad (13)$$

| Width transition area in pixels | Exposure Difference value | Sharpness FFT | Sharpness Laplace | Average pixel shift |
|---|---|---|---|---|
| 5 poor | 2013 | 9,388 | 27,684 | 13,772 |
| 5 rich | 16336 | 28,441 | 41,629 | |
| 25 poor | 1662 | 7,775 | 24,832 | 13,752 |
| 25 rich | 16601 | 28,098 | 37,402 | |
| 50 poor | 1653 | 6,494 | 22,714 | 13,152 |
| 50 rich | 16773 | 27,584 | 33,933 | |
| 150 poor | 2484 | 4,218 | 17,983 | 13,120 |
| 150 rich | 17371 | 25,673 | 24,906 | |

Table 2. Results for determining the best blending parameter

### 4.3 Comparison with Direct Stitching Method

Finally, there will be a comparison of the newly developed feature-based approach with an approach based on the classical direct stitching method. Both algorithms stitch a video sequence with 900 frames. An enlarged cut-out of both results is shown in Fig.11. It can be seen that the newly developed variant is about three times faster and also qualitatively much better. The result of the direct method is much wavier at a similar resolution. This is due to the partially extremely homogeneous ball screw surface. This can lead to inaccuracies in the precise localization of defects. All in all, a superiority of the method presented here has been proven.

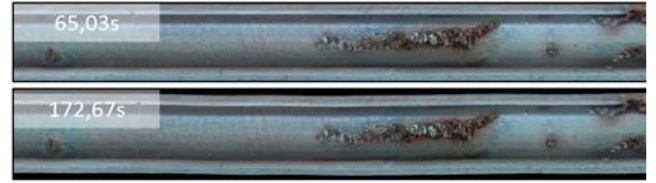

Fig. 11. Result of the newly developed feature-based approach (top) and of a classical stitching method (bottom) and their corresponding processing times

### 4.4 Application in Analysis Pipeline

To verify its functionality, the stitching algorithm is finally integrated into an automatic analysis pipeline. The validation will be performed on a previously presented ball screw. Four recorded test runs are used for this purpose, which show increasing wear. One analysis run consists of stitching, classification, and documentation in the form of a JSON file. The four stitched and classified videos each correspond to one time step ($t1$, $t2$, $t3$, $t4$) and together represent an increasing surface wear of the ball screw.

The stitched and classified results are illustrated in Fig. 13 and show the same spindle section at four different times and also the wear increase therein. The table below shows the number of patches classified as defective according to the JSON files. On the basis of this data, there are now various possibilities for displaying and evaluating the wear development. The choice is entirely up to the user and the objectives. As an example, a representation in the form of a bar chart is presented in Fig. 12. This allows a clear distinction between the developments during individual time steps, and the local allocation of the wear increase, without losing sight of the global development. It becomes apparent how large the total number of defects is at the end, which areas are particularly badly damaged, and when the greatest increase occurred there.

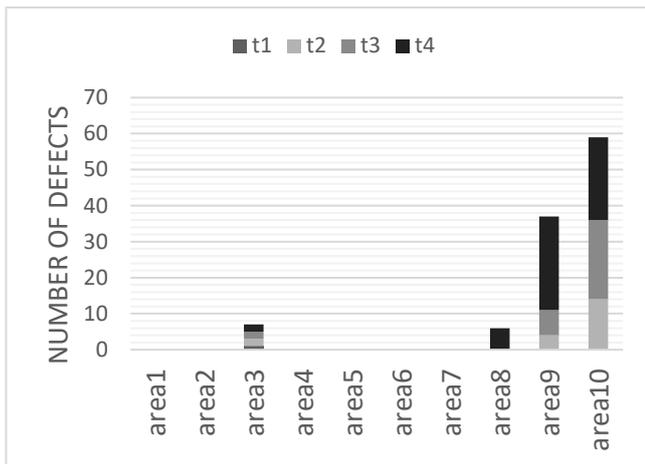

Fig. 12. Bar chart representing the wear development over time

## 5. Conclusion

This paper has presented a new method for stitching rotationally symmetric surfaces. A concrete application was also shown directly. It offers great advantages in the automatic examination of rotationally symmetric surfaces, for example with respect to defects. This was exemplarily presented in the current paper on a ball screw drive spindle but can be easily translated to other applications. In comparison with classical direct stitching methods, it was shown that the algorithm already has a significantly better performance in terms of quality and especially in terms of processing time. The algorithm was written in Python for the first proof of concept. In the further course of the project, the focus of research lies on the speed-up of the used algorithms together with the further improvement of feature extraction algorithms. A closed model to feature-extract, stitch, and detect failures should be investigated.

Another interesting approach that could be pursued in future work is to use the encoded information in the I-, B- and P-Frames of a .h264 video. This makes it possible to directly determine the motion vector between individual frames. For high quality video material, this is an interesting addition to the already developed approach. The determination of the motion vector via a feature extraction could be replaced, which is a good way to further accelerate the stitching process and also to become much more robust in low-featured areas.

## Acknowledgement
This work was supported by the German Research Foundation (DFG) under Grant FL 197/77-1.

## References

Adel, E.; Elmogy, M. & Elbakry, H. (2014), „Image Stitching based on Feature Extraction Techniques: A Survey", *International Journal of Computer Applications*, Nr. 6, S. 1–8.

Balasundaram, M. K. & Ratnam, M. M. (2014), „In-process measurement of surface roughness using machine vision with sub-pixel edge detection in finish turning", *International Journal of Precision Engineering and Manufacturing*, Nr. 11, S. 2239–2249.

Beyerer, J.; Puente León, F. & Frese, C. (2016), *Automatische Sichtprüfung,* Springer Berlin Heidelberg.

Bind, V.; Muduli, P. & Pati, C. (2013), „A Robust Technique for Feature-based Image Mosaicing using Image Fusion".

Brown, M. & Lowe, D. G. (2007), „Automatic Panoramic Image Stitching using Invariant Features", *International Journal of Computer Vision,* S. 59–73.

Chum, O.; Pajdla, T. & Sturm, P. (2005), „The geometric error for homographies", *Computer Vision and Image Understanding*, Nr. 1, S. 86–102.

Dutta, S.; Pal, S. K. & Sen, R. (2016), „Tool Condition Monitoring in Turning by Applying Machine Vision", *Journal of Manufacturing Science and Engineering*, Nr. 5.

Fernández-Robles, L.; Azzopardi, G.; Alegre, E. & Petkov, N. (2016), „Machine-vision-based identification of broken inserts in edge profile milling heads".

Gauglitz, S.; Höllerer, T. & Turk, M. (2011), „Evaluation of Interest Point Detectors and Feature Descriptors for Visual Tracking", *International Journal of Computer Vision*, Nr. 3, S. 335–360.

Jähne, B. (2012), *Digitale Bildverarbeitung und Bildgewinnung,* Springer Vieweg. ISBN: 978-3-642-04951-4.

Kumar, A. (2008), "Computer-Vision-Based Fabric Defect Detection: A Survey".

Kumar, B. M. & Ratnam, M. M. (2015), „Machine vision method for non-contact measurement of surface roughness of a rotating workpiece", *Sensor Review,* S. 10–19.

Massart, D.; Kaufmann, L.; Rousseeuw, P. & Leroy, A. (1986), „Least Median of Squares: A Robust Method for Outlier and Model Error Detection in Regression and Calibration".

Mistry, S. (2016), „Image Stitching using Harris Feature Detection".

Noble, F. (2016), „Comparison of OpcnEV's Feature Detectors and Feature Matchers".

Pech-Pacheco, J.; Cristóbal, G.; Chamorro-Martínez, J. & Fernández-Valdivia, J. (2000), „Ciatom Autofocusing in Brightfield Microscopy: A Comparative Study".

Pertuz, S.; Puig, D. & Garcia, M. (2013), „Analysis of focus measure operators for shape-from-focus".

Qiao, Y.; Li, J. & Tang, Y. (2013), „Improved Harris Sub-pixel Corner Detection Algorithm for Chessboard Image".

Schlagenhauf, T; Hillenbrand, J.; Klee, B. & Fleischer, J. (2019), „Integration von Machine Vision in Kugelgewindespindeln", wt Werkstattstechnik online, 7/8, pp. 605-610.

Simonyan, Karen; Zisserman, Andrew (2014): "Very Deep Convolutional Networks for Large-Scale Image Recognition".

Tsen, T. (2014), „Video stitching Literature Review".

Xiong, Y.; Pulli, K. (2010), "Fast panorama stitching for high-quality panoramic images on mobile phones", IEEE

Zitová, B. & Flusser, J. (2003), „Image registration methods: a survey", *Image and Vision Computing*, Nr. 11, S. 977–1000.

# Appendix

For the sake of readability, the code for the stitching method is available at https://github.com/bra-ti/stitcher

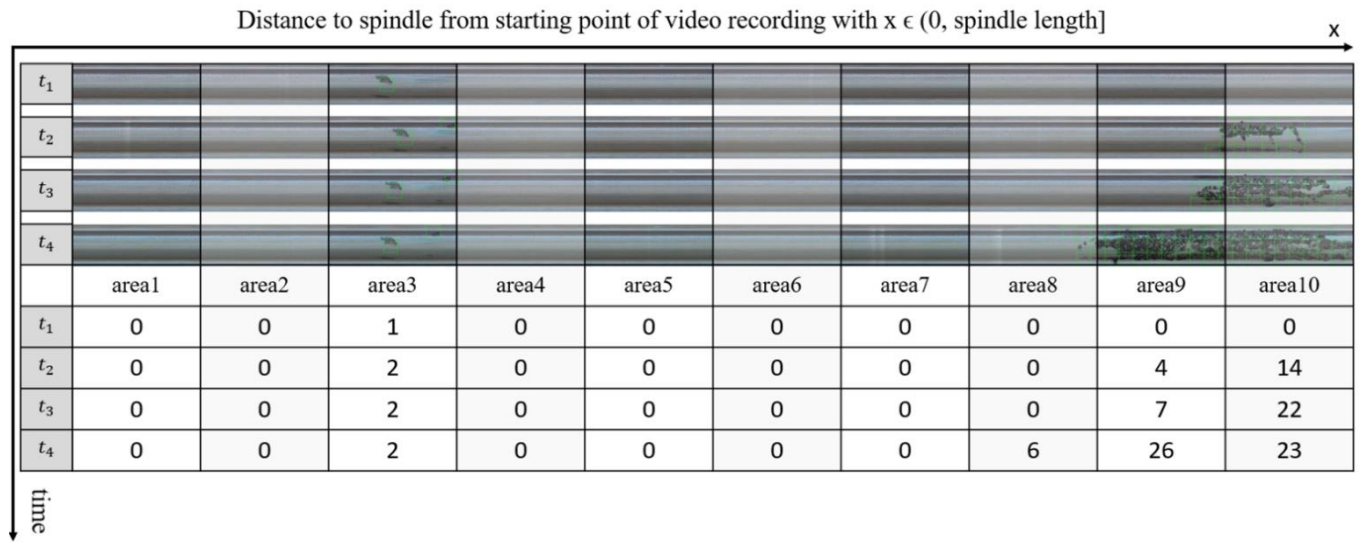

Fig.13. Wear detection over four time steps